\documentclass{article}
\usepackage{spconf,amsmath,graphicx}
\usepackage{hyperref}
\usepackage{enumitem}
\usepackage{float}
\setlist{nosep, leftmargin=14pt}
\usepackage[ruled,vlined,linesnumbered]{algorithm2e}
\usepackage{mwe} 
\usepackage{makecell}
\usepackage{array}
\usepackage{dblfloatfix}

\title{Training a universal instance segmentation network for live cell images of various cell types and imaging modalities}

\twoauthors
  {Tianqi Guo\sthanks{Author to whom any correspondence should be addressed. E-mail: tqguo246@gmail.com}, Yin Wang, Jan P. Allebach}
	{Purdue University\\
	School of Electrical and Computer Engineering\\
	West Lafayette, IN}
  {Luis Solorio}
	{Purdue University\\
	Weldon School of Biomedical Engineering\\
	West Lafayette, IN}
\begin{document}
%
\maketitle
\begin{abstract}
\textit{In this work, we share our recent findings in an attempt to train a universal segmentation network for various cell types and imaging modalities. Our method was built on the generalized U-Net architecture, which allows the evaluation of each component individually. We modified the traditional binary training targets to include three classes for direct instance segmentation. Detailed experiments were performed regarding training schemes, training settings, network backbones, and individual modules on the segmentation performance. Our proposed training scheme draws minibatches in turn from each dataset, and the gradients are accumulated before an optimization step. We found that the key to training a universal network is all-time supervision on all datasets, and it is necessary to sample each dataset in an unbiased way. Our experiments also suggest that there might exist common features to define cell boundaries across cell types and imaging modalities, which could allow application of trained models to totally unseen datasets. A few training tricks can further boost the segmentation performance, including uneven class weights in the cross-entropy loss function, well-designed learning rate scheduler, larger image crops for contextual information, and additional loss terms for unbalanced classes.  We also found that segmentation performance can benefit from group normalization layer and Atrous Spatial Pyramid Pooling (ASPP) module, thanks to their more reliable statistics estimation and improved semantic understanding, respectively. We participated in the 6th Cell Tracking Challenge (CTC) held at IEEE International Symposium on Biomedical Imaging (ISBI) 2021 using one of the developed variants. Our method was evaluated as the best runner up during the initial submission for the primary track, and also secured the 3rd place in an additional round of competition in preparation for the summary publication. All scripts with pre-trained weights are publicly available on GitHub at \url{https://github.com/westgate458/XB-Net}.}

\end{abstract}

\section{Introduction}
\label{sec:intro}

With the recent development in the field of bioimaging, long-duration recordings of live cells with high spatial and temporal resolutions have become much easier to obtain, store, and transmit. The vast availability of such data poses both blessings and curses to this community, as the bottleneck still remains as the proper automation of the image/video processing techniques, to efficiently extract useful information with accuracy comparable to trained human experts. 

The processing of those time-lapse microscopy recordings usually involves the identification, segmentation, and tracking of individual cells. Those are the key steps in order to gain biological insights into the dynamic processes of interest, including cell proliferation and migration. The improvement of those steps can benefit later more advanced analysis, for example the classification of cell types \cite{Zeng2017NeuronalForward}, the quantification of cell growth rate or mobility \cite{Jun2020Fibronectin-ExpressingMetastasis}, and detection of cell-lineage changes \cite{Matula2015CellGraphs}. Current study focuses on the subtask of cell segmentation, which aims to accurately delineate the cell boundaries by performing pixel-wise classification that determines whether each pixel belongs to the foreground cell, or the background.

The traditional methods in this area typically require the design of sophisticated algorithms for a specific cell type recorded by a specific imaging modality. However, the late blooming success of deep-learning based methods for other computer vision tasks \cite{Guo2021RemoteNetworks,Zamir2020CycleISP:Synthesis,Redmon2015YouDetection,He2017MaskR-CNN,Dosovitskiy2015FlowNet:Networks,Xiang2021BoostingSuper-Resolution}, has encouraged some researchers in the cell tracking community to shift away from the traditional route, and inspired rapid advances in a few new frontiers, including the design of novel convolutional neural networks (CNN) architectures \cite{Gadosey2020SD-UNET:Budgets} in the context of cell segmentation, and the improvements of training protocols. For example, the selection of training targets and loss functions \cite{GuerreroPena2020JSegmentation}, etc. This trend is also well reflected in the recent leaderboard changes for the online Cell Tracking Challenge (\url{http://celltrackingchallenge.net/}).

In this study, we followed those recent endeavors, and carried out extensive experiments to address the following four open questions in this area, including:

\begin{enumerate}
\item What is a better way to train a universal cell segmentation network that works for various cell types and imaging modalities?
\item Besides regular hyper-parameter tuning, what other handy tricks should be considered when designing the training protocol? 
\item Is it possible to achieve comparable or even better segmentation performance if we use other backbones to replace the VGG-like encoder in the original U-Net \cite{Ronneberger2015U-Net:Segmentation}?
\item What recently developed CNN modules can be incorporated into the U-Net to boost the segmentation performance?
\end{enumerate}

This paper is structured as follows. In Section \ref{sec:review} we give a brief literature review of related works, followed by an overview of our methodology in Section \ref{sec:method}. The details of the experiments will be presented in Section \ref{sec:exp}, with attempts to shed some light on the four questions above. We also report our participation in the 6th Cell Tracking Challenge in Section \ref{sec:ctc}. The conclusion remarks will be discussed at last in Section \ref{sec:conclusion}.

\section{Related works}
\label{sec:review}

\subsection{Traditional methods for cell segmentation}

Traditional methods for cell segmentation heavily depend on the careful design of consecutive processing steps. A typical processing pipeline usually consists of cell identification based on intensity thresholding, edge detection to draw cell boundaries, morphological operations for edge refinement, and watershed transforms \cite{Meyer2007MorphologicalRevisited} to separate cell clumps. A few studies \cite{Chenouard2014ObjectiveMethods,Ulman2017AnAlgorithms} have provided thorough comparisons of those methods developed for cell segmentation prior to the era of deep learning.

\subsection{Deep-learning methods for semantic segmentation}
Most modern segmentation networks \cite{Ronneberger2015U-Net:Segmentation,Chen2018DeepLab:CRFs,Long2015FullySegmentation,Hariharan2015HypercolumnsLocalization,Noh2015LearningSegmentation,Badrinarayanan2017SegNet:Segmentation} are designed for semantic segmentation where each integer value corresponds to a different semantic category (e.g. tree, floor, pedestrian, etc.). When adapted for cell segmentation, the expected output of such networks is a binary map with all cell pixels labelled as foreground. The distinct cell labels then are typically obtained by an extra post-processing step borrowed from the traditional processing pipelines. 

However, cells from different datasets show distinct morphological characteristics, and even cells within the same dataset usually undergo complex shape deformation during division and migration. Consequently, it is cumbersome to hand-tailor parameters required by those borrowed post-processing techniques, which is an inherited drawback from the traditional methods.

\begin{table*}[!b]
\caption{Datasets used for current study. The two numbers in each cell correspond to two annotated training sequences.}
\centering
\begin{tabular}{ | m{3.2cm} | m{0.75cm}| m{3.6cm} | m{2.5cm} | m{2.25cm}| m{2.5cm} | } 
\Xhline{4\arrayrulewidth}
\makecell{\textbf{Dataset}} & \makecell{\textbf{Type}} & \makecell{\textbf{Resolution}}& \makecell{\textbf{Total Frames$^*$}} & \makecell{\textbf{GT Frames}} & \makecell{\textbf{ST Frames}} \\ 
\Xhline{4\arrayrulewidth}
\makecell{BF-C2DL-HSC} & \makecell{2D} & \makecell{1010$\times$1010, 1010$\times$1010}  & \makecell{1764, 1764} & \makecell{49, 8} & \makecell{1764, 1764} \\
\hline 
\makecell{BF-C2DL-MuSC} & \makecell{2D} & \makecell{1070$\times$1036, 1070$\times$1036}  & \makecell{1376, 1376} & \makecell{50, 50} & \makecell{1376, 1376} \\
\hline 
\makecell{DIC-C2DH-HeLa} & \makecell{2D} & \makecell{512$\times$512, 512$\times$512}  & \makecell{84, 84} & \makecell{9, 9} & \makecell{84, 84} \\
\hline 
\makecell{Fluo-C2DL-MSC} & \makecell{2D} & \makecell{992$\times$832, 1200$\times$782}  & \makecell{48, 8} & \makecell{18, 33} & \makecell{48, 48} \\
\hline 
\makecell{Fluo-N2DH-GOWT1} & \makecell{2D} & \makecell{1024$\times$1024, 1024$\times$1024}  & \makecell{92, 92} & \makecell{30, 20} & \makecell{92, 92} \\
\hline 
\makecell{Fluo-N2DL-HeLa} & \makecell{2D} & \makecell{1100$\times$700, 1100$\times$700}  & \makecell{92, 92} & \makecell{28, 8} & \makecell{92, 92} \\
\hline 
\makecell{PhC-C2DH-U373} & \makecell{2D} & \makecell{696$\times$520, 696$\times$520}  & \makecell{115, 115} & \makecell{15, 19} & \makecell{115, 115} \\
\hline 
\makecell{PhC-C2DL-PSC} & \makecell{2D} & \makecell{720$\times$576, 720$\times$576}  & \makecell{300, 300} & \makecell{2, 2} & \makecell{300, 300} \\
\Xhline{4\arrayrulewidth}
\makecell{Fluo-C3DH-A549} & \makecell{3D} & \makecell{350$\times$300, 400$\times$270}  & \makecell{29$\times$30, 28$\times$30} & \makecell{29$\times$15, 28$\times$15} & \makecell{29$\times$30, 28$\times$30} \\
\hline 
\makecell{Fluo-C3DH-H157} & \makecell{3D} & \makecell{992$\times$832, 992$\times$832}  & \makecell{35$\times$60, 80$\times$60} & \makecell{33, 40} & \makecell{35$\times$60, 80$\times$60} \\
\hline 
\makecell{Fluo-C3DL-MDA231} & \makecell{3D} & \makecell{512$\times$512, 512$\times$512}  & \makecell{30$\times$12, 30$\times$12} & \makecell{17, 13} & \makecell{30$\times$12, 30$\times$12} \\
\hline 
\makecell{Fluo-N3DH-CE} & \makecell{3D} & \makecell{708$\times$512, 712$\times$512}  & \makecell{35$\times$250, 31$\times$250} & \makecell{5, 5} & \makecell{35$\times$195, 31$\times$190} \\
\hline 
\makecell{Fluo-N3DH-CHO} & \makecell{3D} & \makecell{512$\times$443, 512$\times$443}  & \makecell{5$\times$92, 5$\times$92} & \makecell{19, 24} & \makecell{5$\times$92, 5$\times$92} \\
\Xhline{4\arrayrulewidth}
\end{tabular}
\label{tb:datasets}
\\\vspace*{2mm} \footnotesize{$^*$ For 3D datasets: (number of slices in a volume stack) $\times$ (time steps)}\\
\end{table*} 

\subsection{Deep-learning methods for instance segmentation}
Cell segmentation is by nature an instance-wise segmentation problem, where the desired output is a discrete map where all pixels with the same integer value corresponding to the same cell object.

Instance segmentation can be achieved by joint-training with an object detection network (e.g. as in Mask R-CNN \cite{He2017MaskR-CNN}), which first predicts a regressed bounding box for each detected object, then the foreground-background binary segmentation map for this particular object is predicted within the proposed subregion. This approach has proven successful in other instance segmentation tasks including scene parsing \cite{Kirillov2019PanopticSegmentation}, motion recognition \cite{Bu2021BedImages}, and ROI filtering \cite{Guo2021RemoteNetworks}. However, for fine-grained segmentation tasks including the problem at hand, this two-step approach requires separate annotation of bounding boxes for each cell, and the accuracy of segmentation heavily depends on the reliability of the preceding bounding box localization step.

Recently, there are some noticeable efforts for direct instance segmentation. Bai \textit{et al.} built upon classical watershed transformation for segmentation, and used deep neural networks to predict unit vectors at each foreground pixel pointing directly away from the nearest boundary \cite{Bai2017DeepSegmentation}.  Based on this unit vector map, watershed energy levels at each pixel are calculated and used for splitting contacting objects apart. In a more recent work, Liang \textit{et al.} modified the DeepLab network \cite{Chen2018DeepLab:CRFs} that was designed for semantic segmentation, to predict two continuous maps where at each pixel the map values correspond to the relative coordinates of the instance bounding box that pixel belongs to \cite{Liang2018Proposal-FreeSegmentation}. Then instance segmentation could be achieved by off-the-shelf clustering algorithms operating on the predicted maps to identify pixels that belong to the same object bounding box. In a concurrent work also dedicated to cell segmentation, Pena \textit{et al.} modified the binary training target for U-Net to contain four different semantic categories: background, cell, touching, and gap \cite{GuerreroPena2020JSegmentation}. When pixels that correspond to background-cell boundaries and cell-cell boundaries are predicted as a new semantic group, the main bodies of the cells are naturally separated as distinct instances.

\begin{figure}[!hb]
  \includegraphics[width=1.0\columnwidth]{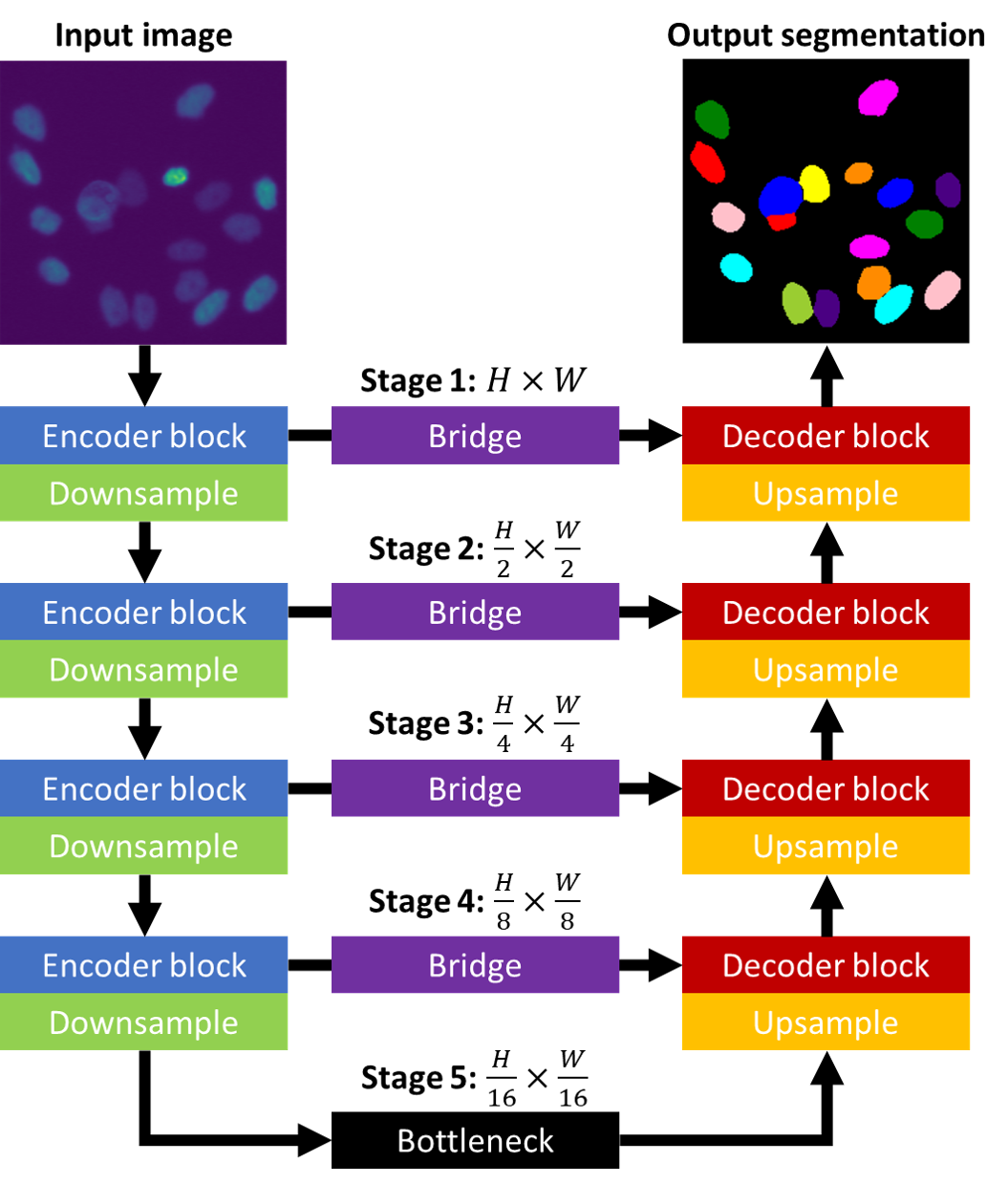}
  \vspace*{-8mm}
  \caption{Generalized network architecture for the U-Net. Final classification layer and optional modules (e.g. entry flow blocks) are omitted for simplicity.}
  \label{Figure: unetArc}
\end{figure}

\subsection{Generalized U-Net}
Since its introduction, U-Net \cite{Ronneberger2015U-Net:Segmentation,Falk2018U-Net:Morphometry} has become the most popular network architecture for segmentation of biomedical images \cite{GuerreroPena2020JSegmentation,Chen2020ResidualImages,Gadosey2020SD-UNET:Budgets}. Like other network architectures used for dense prediction tasks \cite{Dosovitskiy2015FlowNet:Networks,Noh2015LearningSegmentation,Ilg2017FlowNetNetworks,Badrinarayanan2017SegNet:Segmentation}, U-Net has a two-part multi-stage encoder-decoder structure as shown in Figure \ref{Figure: unetArc}, joined by the bottleneck pathway as well as several bridges for skip-connection at each stage.

The encoder part has alternating encoder blocks and downsampling layers. The encoder block extracts abstract features from the image and may contain consecutive convolutional layers, normalization layers (e.g. batch normalization \cite{Ioffe2015BatchShift}, or group normalization \cite{Wu2018GroupNormalization}, etc.), and activation functions (e.g. ReLU \cite{Nair2010RectifiedMachines}, or Leaky ReLU \cite{He2015DelvingClassification}, etc.). The downsampling layers (e.g.  max pooling, or convolution with strides, etc.) on the other hand reduce the spatial dimensions of the feature maps, and effectively broaden the receptive fields of convolutional filters on the input image.  Therefore, semantic features can be learned with richer contextual information in deeper stages. The original U-Net uses VGG-like \cite{Simonyan2015VeryRecognition} encoder blocks, and $2 \times2$ max pooling layers for downsampling.

The decoder part similarly has alternating upsamling layers and encoder blocks. The upsampling layers (e.g. bilinear interpolation, unpooling \cite{Noh2015LearningSegmentation}, deconvolution \cite{Zeiler2010DeconvolutionalNetworks}, or pixelshuffle \cite{Shi2016Real-TimeNetwork}) restore the spatial dimensions of the feature maps. Consequently, fine details of the boundaries between semantic groups can be recovered via reconstruction in the decoder blocks, which may again consist of consecutive convolutional layers, normalization layers, and activation functions. The original U-Net uses the same VGG-like  blocks in decoder, and $2 \times2$ deconvolutional layers for upsampling.

The encoder and the decoder are connected by the bottleneck pathway as well as the bridges. The original U-Net uses a VGG-block at the bottleneck, which has the smallest feature map dimensions and largest number of feature channels, where semantic information is richly encoded. Another signature component in the U-Net architecture is the bridges that forward the feature maps from the encoder towards the decoder, which helps fine-grained localization in segmentation tasks \cite{Hariharan2015HypercolumnsLocalization}. In almost any modern CNN, the fine feature maps from the shallow layers contain appearance information of the local pixels, while the coarse feature maps from the deep layers have rich semantic information regarding the class of the object in the local region \cite{Long2015FullySegmentation}. The original U-Net uses simple identity mapping with center cropping as the bridges since the number of channels in encoder and decoder are equal at each stage. Feature maps in decoder after upsampling are concatenated with feature maps of the same size from the encoder via those bridges. Moreover, those bridges also serve as skip connections that ease the training of deep networks by providing additional gradient flow \cite{He2016DeepRecognition}.

\section{Method}
\label{sec:method}

\subsection{Datasets used for study}
Eight 2D datasets and five 3D datasets of time-lapse live cell images are obtained from the Cell Tracking Challenge website (\url{http://celltrackingchallenge.net/}). Each dataset has four image sequences released to the participants: 1) Two of those sequences are the training sets, which have two types of annotations: ground truth (GT) segmentation and tracking annotated by human experts, silver truth (ST) segmentation as a result of majority voting from previous online submissions. The image resolutions, number of frames, number of available GT and ST annotations for those two sequences are listed in Table \ref{tb:datasets}. 2) The other two image sequences form the challenge (test) set, the segmentation and tracking of which are kept secret and used to evaluate the performance of the participant algorithms.

Images from those datasets were captured by various imaging modalities with different resolutions, formats and dynamic ranges. For example, bright field (BF), phase contrast (PhC), and differential interference contrast (DIC) microscopes are usually coupled with a 8-bit camera, while images captured on fluorescence (Fluo) microscopes can be saved in a 16-bit format. Sometimes even when the same image format is used, the intensity ranges can vary from dataset to dataset. For example, both datasets Fluo-C3DH-A549 and Fluo-C2DL-MSC use a 16-bit image format, however in the first dataset the intensities range from 359 to 4810, with little overlap with the intensity range from 2816 to 65280 in the second dataset. 

\begin{figure}[H]
  \includegraphics[width=1.0\columnwidth]{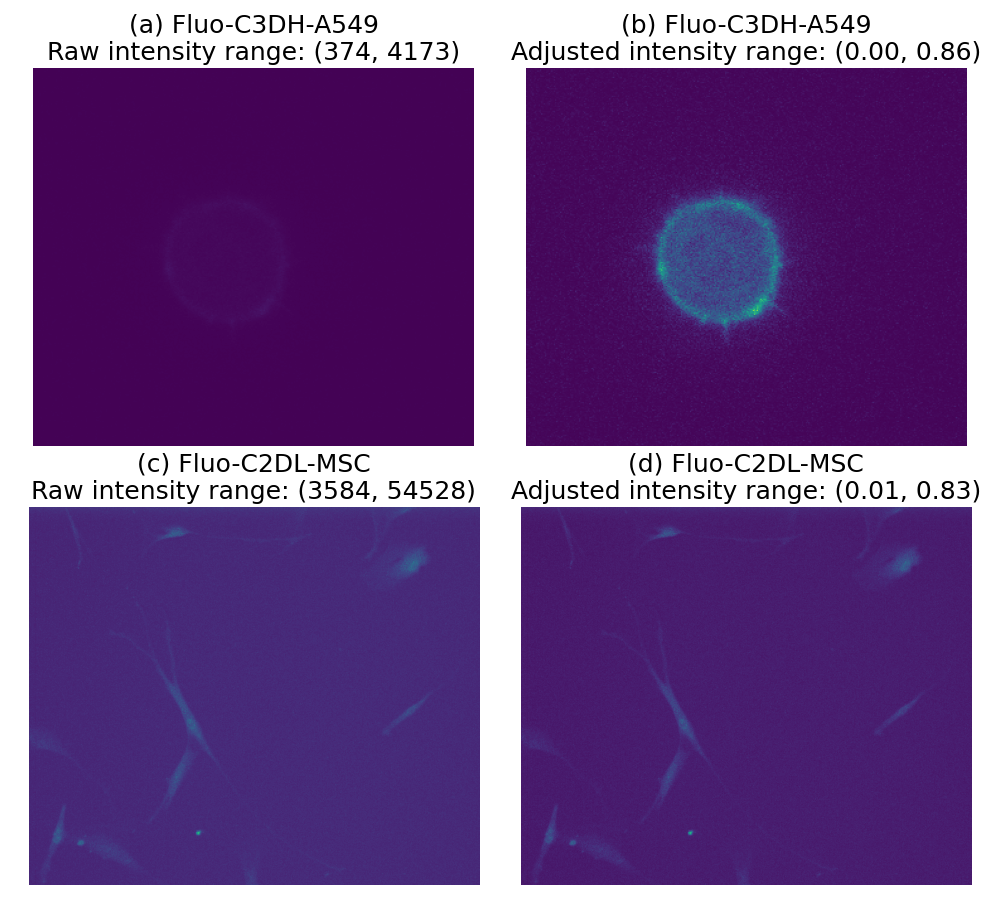}
  \vspace*{-9mm}
  \caption{Examples of image preprocessing on two datasets with 16-bit image format. Rendered in false color for better visualization. Best viewed in color.}
  \label{Figure: imgPre}
\end{figure}

\subsection{Image pre-processing}
\label{sec:imgPre}
In order to train a universal network and apply the same training protocol across 2D and 3D datasets, we convert the 3D image stacks to planar images such that the same 2D CNN architecture can work for all data configurations, as later explained in Section \ref{sec:ctc_intro}. We also normalize image sequences (01, 02) for each training set by their corresponding maximum and minimum intensity values over the entire video duration. Example images before and after intensity normalization are shown in Figure \ref{Figure: imgPre}. We then combine the two image sequences for each dataset as a single set for training.  During inference on test set, the images are normalized in a similar fashion.

\begin{figure*}[btp]
  \includegraphics[width=2.0\columnwidth]{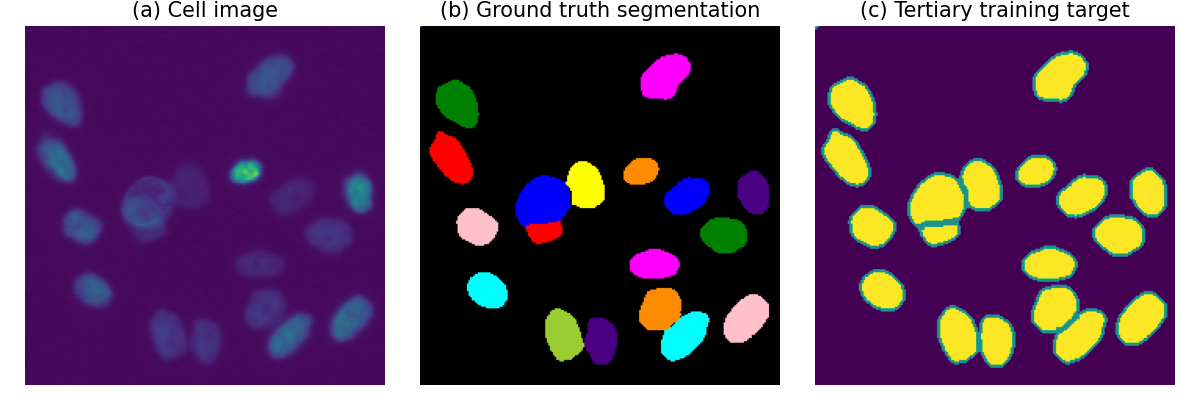}
  \vspace*{-5mm}
  \caption{Examples from the Fluo-N2DL-HeLa dataset: (a) Original cell image. (b) Provided ground truth instance segmentation map. Rendered in false color for better visualization. Each color corresponds to a cell instance. (c) Converted tertiary training map for pixel-wise classification. Three colors indicate semantic groups of background, cell, and boundary. Best viewed in color.}
  \label{Figure: trainingTarget}
\end{figure*}

\subsection{Training target}
\label{sec:trainTarget}
\newcommand{\var}{\texttt}

In current work, we follow a similar logic as in \cite{GuerreroPena2020JSegmentation}, but only three semantic groups are used: background, cell, and boundary. The conversion from the ground truth instance-wise segmentation map to the tertiary training target (with $\var{label\_background}$, $\var{label\_boundary}$, and $\var{label\_cell}$) is described in Algorithm \ref{alg:training_target}. More specifically, the following operations are performed for each cell labelled in the ground truth segmentation map:

\begin{algorithm}[h]
\SetAlgoLined
\KwIn{instance labels $\var{seg}_{GT}$}
\KwOut{training target $\var{map}_{3cls}$}
$\var{map}_{3cls}$ $\gets$ initialized as $\var{label\_background}$\;
background $\gets$ ($\var{seg}_{GT}$\ == $\var{label\_background}$)\;
cell\_list $\gets$ unique instance labels in $\var{seg}_{GT}$\;
\For{${cell\_id}\gets$ cell\_list}{
    mask $\gets$ ($\var{seg}_{GT}$ == ${cell\_id}$)\;
    dilated\_mask $\gets$ dilation on mask\;
    dilated\_mask $\gets$ dilated\_mask \& background\;
    eroded\_mask $\gets$ erosion on mask\;
    boundary\_mask $\gets$ mask - eroded\_mask\;
    $\var{map}_{3cls}$[dilated\_mask] $\gets$ $\var{label\_boundary}$\;
    $\var{map}_{3cls}$[mask] $\gets$ $\var{label\_cell}$ \;
    \For{${pixel}\gets$ boundary\_mask}{
        \uIf{${pixel}$ is next to another cell in $\var{seg}_{GT}$}{
            $\var{map}_{3cls}$[${pixel}$] $\gets$ $\var{label\_boundary}$\;
        }
    }
}
 \caption{Obtain tertiary training target}
 \label{alg:training_target}
\end{algorithm}

\begin{enumerate}
\item On Line 6-7, a dilation operation (2 iterations for 2D datasets, and 5 iterations for 3D datasets) on the binary segmentation map for this cell gives the dilated mask. The intersection between the dilated mask and the ground truth background gives the boundary pixels between cell and background.
\item On Line 8-9, an erosion operation (2 iterations for all datasets) on the binary segmentation map for this cell gives the eroded mask. The pixels that belong to the original binary segmentation map, but do not belong to this eroded mask are potential boundary pixels that demand further examination.
\item On Line 14, among those candidate pixels the ones that are in direct contact with, or within a certain distance away from, another cell will be labelled as the boundary pixels between contacting cell instances.
\end{enumerate}

Examples for the cell image, original ground truth, and modified tertiary training target are shown in Figure \ref{Figure: trainingTarget}.

\subsection{Implementation details}
\label{sec: implementation}

We apply standard data augmentation on the (normalized image, tertiary segmentation map) pair, including random square cropping, random horizontal/vertical flipping, random 90\textdegree{} rotations. We also pad around the square crops with 8-pixel (16-pixel in the case of a 6-stage U-Net) reflections on each side.

All networks and training scripts are implemented in PyTorch \cite{Paszke2019PyTorch:Library}. For loss function, the off-the-shelf PyTorch implementation of the cross-entropy loss (as a combination of log of softmax loss and negative log likelihood loss) is used unless otherwise stated. The default weights for background, boundary, and cell classes are set to (1, 10, 5) to address the issue of unbalanced classes.  The loss functions are optimized using the AdamW optimizer \cite{Loshchilov2019DecoupledRegularization} with default settings. The learning rate is kept constant at $10^{-4}$ unless a scheduler is used.

Models reported in this work were trained on a single GPU (randomly assigned Tesla P100-PCIE-16GB, Tesla V100-PCIE-16GB, Tesla V100-PCIE-32GB, or A100-PCIE-40GB) available on the Purdue Gilbreth community cluster \cite{McCartney2014}. Measurements of model parameters and multiply-add operations are estimated by the Python package ptflops v0.6.6 (\url{https://pypi.org/project/ptflops/}).  The GPU memory consumption and time spent for each training iteration are estimated on the slowest GPU available (Tesla P100-PCIE-16GB).

For all evaluations on the validation sets, we use the Intersection over Union (IoU) metric to quantify the segmentation performance. The output segmentation maps are written onto hard drives, and then analyzed by the official IoU evaluation executable provided by the CTC organizer.

\section{Experiment}
\label{sec:exp}

In this section, we report extensive experiments to address the four questions introduced in Section \ref{sec:intro}. Firstly, we explore what training schemes give the best universal model that works well on all 13 datasets. Secondly, we further test a few training settings to see their effects on the segmentation performance. Next, we modify the U-Net architecture and replace its encoder with other popular backbones. Finally, we evaluate the segmentation performance after replacing some modules in U-Net with their alternatives.

For all trainings carried out in this section, we use all the frames with ST annotation as the training set, and all the frames with GT annotation as the validation set. This is the same splitting rule we used for the allGT+allST data configuration in Section \ref{sec:ctc}.  For each variant tested, the model training is repeated 5 times to report the mean and standard deviation for the best IoUs and their corresponding iteration numbers.

\subsection{Effect of training scheme}
\label{sec:trainScheme}
\subsubsection{Experiment details}

Experiments in this section aim to train a universal network for all 13 datasets. More specifically, we trained the original U-Net with the schemes below. Except for the scheme Mix, every dataset has its own dataloader. And for all training schemes we perform one optimization step after each minibatch, except for the scheme Acc. For schemes Seq, Fix, and Acc, the datasets are arranged in the same order as listed in Table \ref{tb:datasets}. 

\begin{figure*}[!t]
  \includegraphics[width=2.0\columnwidth]{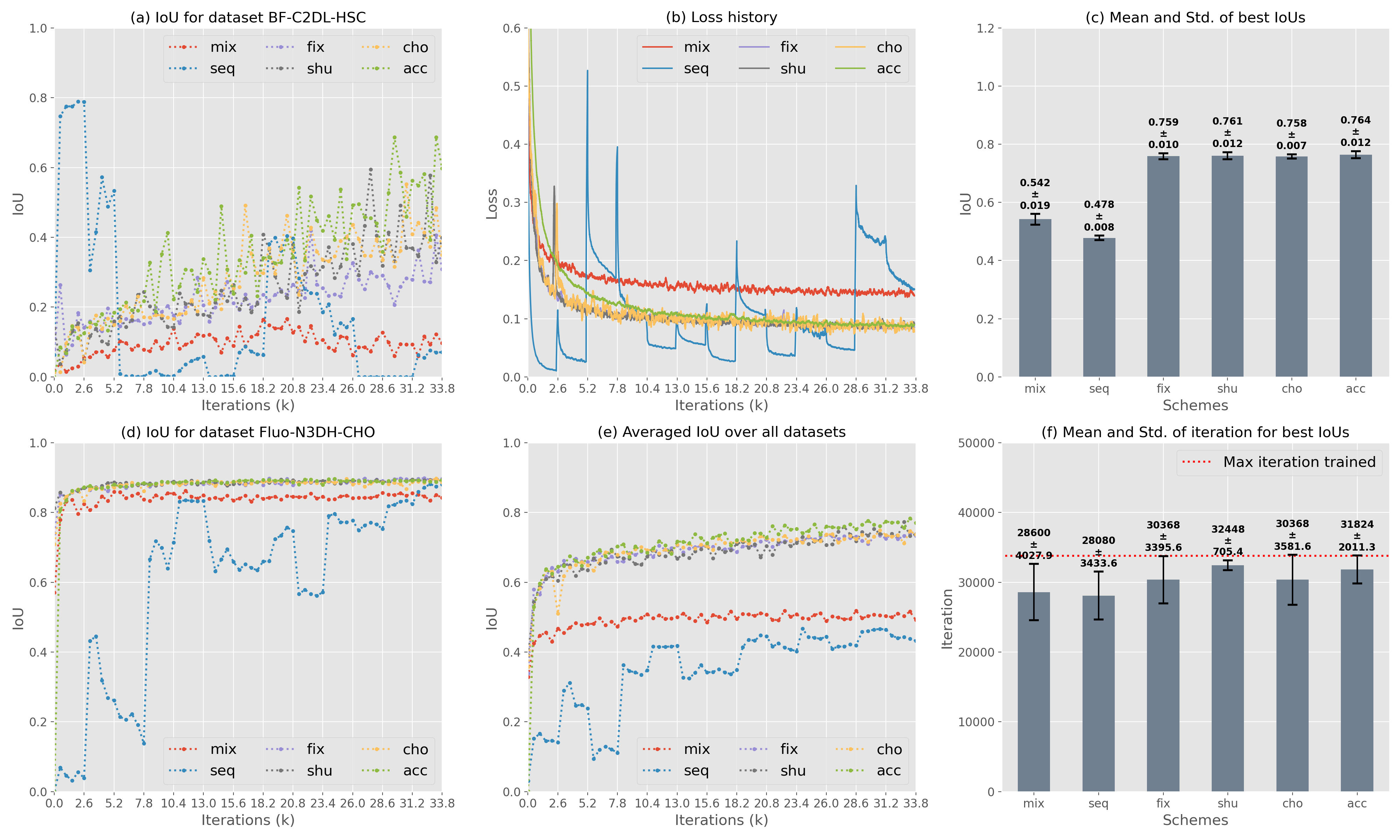}
  \vspace*{-1mm}
  \caption{Experiment results on different training schemes. The vertical white lines in (a-b) and (d-e) indicate the iteration where the model starts to be trained on a new dataset for scheme Seq. The loss history in (b) is smoothed by a 130-point moving average window for better visual presentation.}
  \label{Figure: sixSchemes}
\end{figure*}

\begin{enumerate} 

\item \textbf{Mix:} Mix all 13 datasets together as a single dataloader, and at each iteration draw a minibatch consisting of images from the combined dataset randomly with equal probability.
\item \textbf{Seq:} Train on each dataloader sequentially in the pre-defined fixed order. At each stage, the network only sees training images from one dataset.
\item \textbf{Fix:} Minibatches are drawn in turn from each dataloader, in the pre-defined fixed order.
\item \textbf{Shu:} Minibatches are drawn in turn from each dataloader. The order of the dataloaders is shuffled once 13 minibatches are drawn. 
\item \textbf{Cho:} At each iteration, randomly choose one dataloader to draw a minibatch. Seeing equal number of training examples from each dataset is therefore not explicitly enforced.
\item \textbf{Acc:} Minibatches are drawn in turn from each dataloader, in the pre-defined fixed order. However, gradients are accumulated from  13 minibatches before making an optimization step.
\end{enumerate}

\begin{algorithm}[h]
\caption{Training procedure for scheme Acc}
\label{alg:training_procedure}
\SetAlgoLined
\KwIn{$\var{dataloaders}_{train}$ and $\var{dataloaders}_{valid}$}
\KwOut{Optimized $\var{model}_{best}$.}
optimizer $\gets$ initialization\;
\While {$itr < itr_{max}$}{
    optimizer $\gets$ zero gradients\;
    \For{$dataloader\gets$ $\var{dataloaders}_{train}$}{
        $itr \gets itr + 1$\;
        images, targets $\gets$ next batch in $dataloader$\;
        predictions $\gets$ $\var{model}$ inference on images\;
        losses $\gets$ criteria on predictions and targets\;
        optimizer $\gets$ back-propagation of losses\;
        }
    optimizer step\;
    \uIf{$itr$ is for validation}{
        IoUs $\gets$ evaluation on $\var{dataloaders}_{valid}$ \;
        \uIf{new best $IoU_{mean}$}{
            $\var{model}_{best} \gets \var{model}$\;
            }
        }
    }
\end{algorithm}

An example training procedure for scheme Acc is shown in Algorithm \ref{alg:training_procedure}. Between Line 4 and Line 10 minibatches are drawn from all 13 dataloaders in turns. For each minibatch the losses are back propagated to model parameters and the gradients are accumulated in-place. The gradient descent on model parameters is then performed on Line 11 as a result of the accumulated gradient over the minibatches from all 13 datasets. Other training schemes can be easily obtained by minor modifications. For example, putting Line 3 and Line 11 within loop between 4 to 10 will yield scheme Fix, and shuffling $\var{dataloaders}_{train}$ on Line 4 will yield scheme Shu. Theoretically, the last 4 training schemes should yield models of equivalent performances when enough training iterations are performed.

To make a fair comparison between training schemes, the same computational budget is used for all training trials. All dataloaders are set to have a batchsize of 24, with each batch consisting of randomly-cropped images of $256 \times 256$ pixels. We empirically found that 33,800 iterations of training is adequate to show the difference between schemes, and that corresponds to equally 2.6k minibatches drawn from each dataset, except for the scheme Mix. The performance of the models is evaluated every 520 iterations, and based on the mean IoU on all 13 validation sets we select the best models to report.

\subsubsection{Experiment results}

Histories of the loss function and the averaged IoU over all datasets are shown in Figure \ref{Figure: sixSchemes} (b) and (e) for representative runs.  It is clear that schemes Fix, Shu, Cho, and Acc show similar training history, and further training is likely to improve the overall performance on all datasets. One minor exception is the scheme Acc, which shows more stable loss history, despite the fact that the loss function is slower to decrease at the initial training stage. This is probably because with accumulation of gradients, the optimization step is performed way less frequent than other schemes. However, since each optimization step is based on the accumulated (not averaged) gradient, the total gradients learned should be equivalent, and the optimization always advances towards the direction where improvements are seen on all 13 datasets.

On the other hand, both training schemes Mix and Seq show inferior results to the other four variants. The Scheme Mix converges to a lower averaged IoU, as datasets with fewer training examples are under-represented in the sampling process. One common way to mitigate this is to assign unequal weights (e.g. reciprocal of number of training examples) to each dataset in the loss function, or to manually adjust the probability of images being sampled from each dataset. 

The scheme Seq shows the worst performance, with unstable loss function and IoU history. Nevertheless, this training scheme shows some interesting characteristics. Representative IoU history curves for the first (`BF-C2DL-HSC') and last (`Fluo-N3DH-CHO') datasets are shown in Figure \ref{Figure: sixSchemes} (a) and (d) for comparison between different training schemes. It is worthwhile noting that the performance on the validation set for `BF-C2DL-HSC' reaches the highest at Iteration 2.6k, after which the model continues to be trained on another dataset (`BF-C2DL-MuSC'). The IoU for `BF-C2DL-HSC' drops significantly while the model adapts to other datasets, which shows the necessity for all-time supervision on all datasets. It is also intriguing to see that the model learns to predict correct segmentation masks for dataset `Fluo-N3DH-CHO' much earlier (e.g. at Iteration 10.4k while training on `Fluo-N2DH-GOWT1') than it starts to see training examples from this dataset at Iteration 31.2k. This reveals the fact that there exist shared features among different datasets to define class boundaries, and the trained model has the potential to be applied to totally unseen imaging modalities or cell types, and provide reasonable segmentation predictions.

The statistics for multiple runs of each training scheme are shown in Figure \ref{Figure: sixSchemes} (c) and (f). The means and standard deviations for the best IoUs on validation sets during training are shown in Figure \ref{Figure: sixSchemes} (c). Training schemes Fix, Shu, Cho, and Acc show equivalent segmentation performance for the training iterations performed. It is not expected that one should have superior IoU than the other three even after training has converged. Training schemes Mix and Seq are easier to implement, while showing much poorer segmentation accuracy. As for the number of iterations to achieve the best IoUs, there are large spreads for almost all the training schemes as shown in Figure \ref{Figure: sixSchemes} (f). Careful selection of initial training rate, adopting better learning rate scheduler, or simply performing more training iterations may help reduce this spread. 

\subsection{Effect of training settings}
\label{sec:trainSettings}
\subsubsection{Experiment details}
Experiments in this section were carried out by training the original U-Net with a few modifications of the training settings to test their effects on the segmentation performance. We adopt the training scheme Acc from last section, due to its more stable loss function profile. More specifically, we trained the universal network with the training setting variants described below: 

\begin{figure*}[!b]
  \includegraphics[width=2.0\columnwidth]{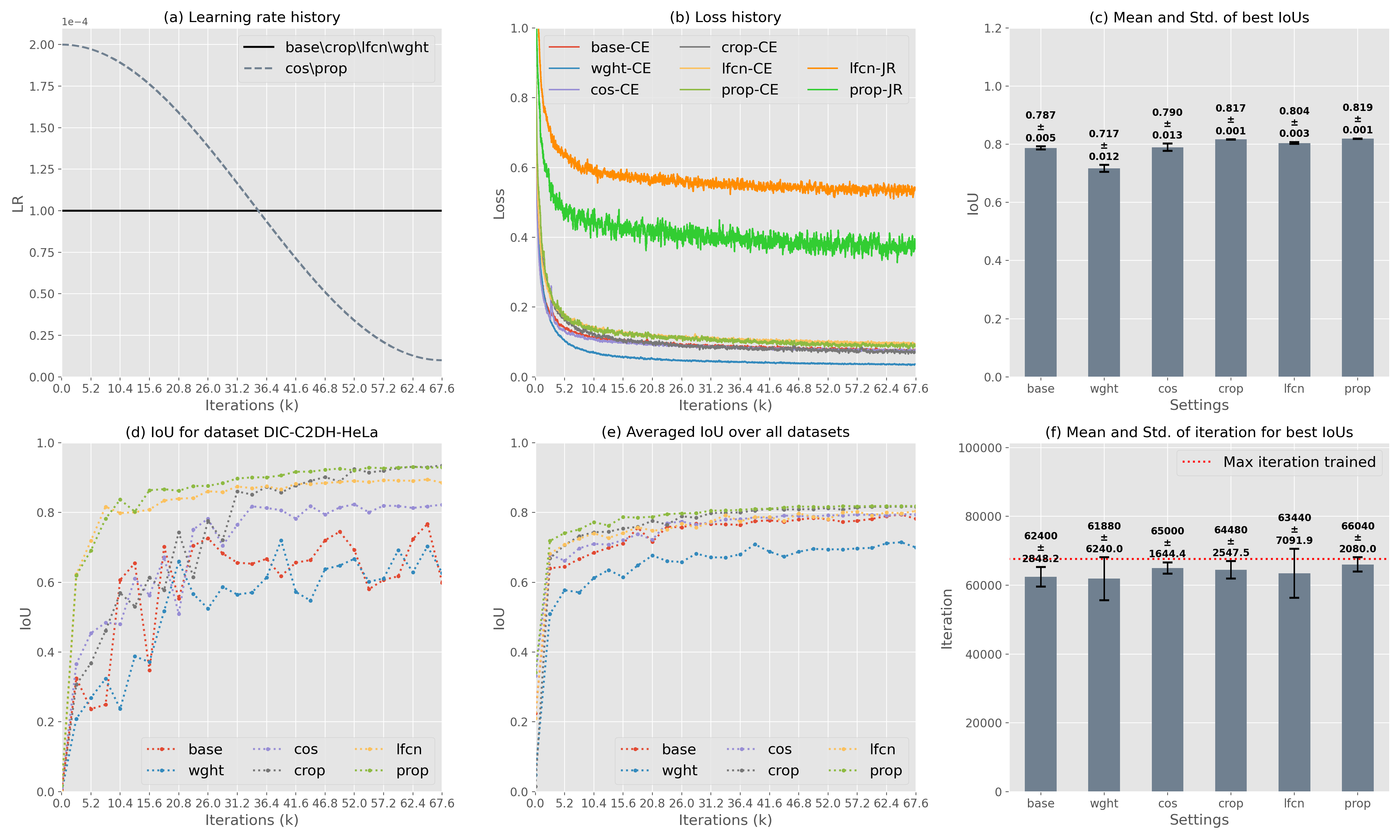}
  \vspace*{-1mm}
  \caption{Results of experiment on different training settings. The loss history in (b) is smoothed by a 130-point moving average window for better visual presentation. The curves indicated by `CE' are cross-entropy loss, while the `JR' ones correspond to J-regularization loss.}
  \label{Figure: trainingSetting}
\end{figure*}

\begin{enumerate} 

\item \textbf{Base:} The same training setting as the previous scheme Acc, where a single cross-entropy loss with weights (1, 10, 5) for the 3 semantic classes, constant learning rate of $1\times10^{-4}$, and random crops of 256$\times$256 pixels are used.

\item \textbf{Wght:} We assign equal weights of (5, 5, 5) to the 3 semantic classes in the cross-entropy loss.

\item \textbf{Cos:} The cosine annealing learning rate scheduler \cite{Loshchilov2016SGDR:Restarts} without the warm restarts is used. The max and min learning rates are set to $2\times10^{-4}$ and $1\times10^{-6}$ as shown in Figure \ref{Figure: sixSchemes} (a), maintaining the same effective amount of gradients learned as other variants.

\item \textbf{Crop:} A random crop of 512$\times$512 pixels is used for datasets whose shorter image side length is at least 512 pixels, while datasets with smaller images still use random crops of 256$\times$256 pixels.

\item \textbf{Lfcn:} Besides the cross-entropy loss, we use an additional J-regularization loss \cite{GuerreroPena2020JSegmentation} to deal with unbalanced classes. The summation of the two losses with equal weights are jointly optimized.

\item \textbf{Prop:} Our proposed training settings, with unequal weights (1,10,5) in the cross-entropy loss, additional J-regularization loss, 512$\times$512 random crops, and cosine annealing learning rate scheduler.

\end{enumerate}

All trainings were carried out for 67,600 iterations, and the models were evaluated on validation sets for every 2,600 iterations.

\subsubsection{Experiment results}
In the ground truth segmentation maps, the class background has way more pixels than the other two classes, and the class boundary has the least amount of training examples. Due to the unbalanced classes, when equal weights were assigned in the cross-entropy loss, the model is more biased towards the background class, which results in a much lower IoU score when compared to the baseline as shown in Figure \ref{Figure: trainingSetting} (c). Please note that, this Wght variant actually shows the lowest loss values when compared with other variants as shown in Figure \ref{Figure: trainingSetting} (b).

Although the final cross-entropy loss of the Lfcn variant is slightly higher than the four cross-entropy-only variants as shown in Figure \ref{Figure: trainingSetting} (b), its final segmentation performance is actually better than the baseline thanks to the joint optimization. Moreover, the J-regularization loss also helps the IoU score on some datasets improve much faster than the four cross-entropy-only variants, taking the dataset DIC-C2DH-HeLa as an example as shown in Figure \ref{Figure: trainingSetting} (d). 

Due to a higher initial learning rate in the Cos variant, the initial convergence speed is accelerated as reflected by a faster rising of the averaged IoU score when compared with the baseline as shown in Figure \ref{Figure: trainingSetting} (e). However, as the training continues it only slightly improves the final segmentation performance over the baseline as shown in Figure \ref{Figure: trainingSetting} (c). 

On the contrary, we found larger training patches lead to a much more profound improvements in final IoU. We suspect this is due to the availability of more contextual information, which is critical when the model needs to adapt to datasets of different cell types and imaging modalities. In other words, when larger windows are visible by the network, global features may provide some hints to help distinguish between datasets, and accordingly the model can adopt different rules when defining the classes. As a matter of fact, images from different datasets do show distinct visual characteristics (e.g. overall brightness and contrast, cell density and appearances, etc.), which experienced human experts can rely on to identify the cell types and imaging modalities by just a glance.

Finally, when we combine the advantages of the above variants, the Prop variant shows superior training profile when compared with all other variants as shown in Figure \ref{Figure: trainingSetting} (d-e). It has the fastest initial convergence speed, which might be important when limited computational budget is affordable and the model is likely to be under-trained. It also achieves the best final IoU score with much smaller standard deviation across runs as shown in Figure \ref{Figure: trainingSetting} (c).  Moreover, the iteration when the best IoU occurs is much later than the other variants as shown in Figure \ref{Figure: trainingSetting} (f), and further training is likely to increase the IoU scores even more.

\begin{figure*}[!ht]
  \includegraphics[width=2.0\columnwidth]{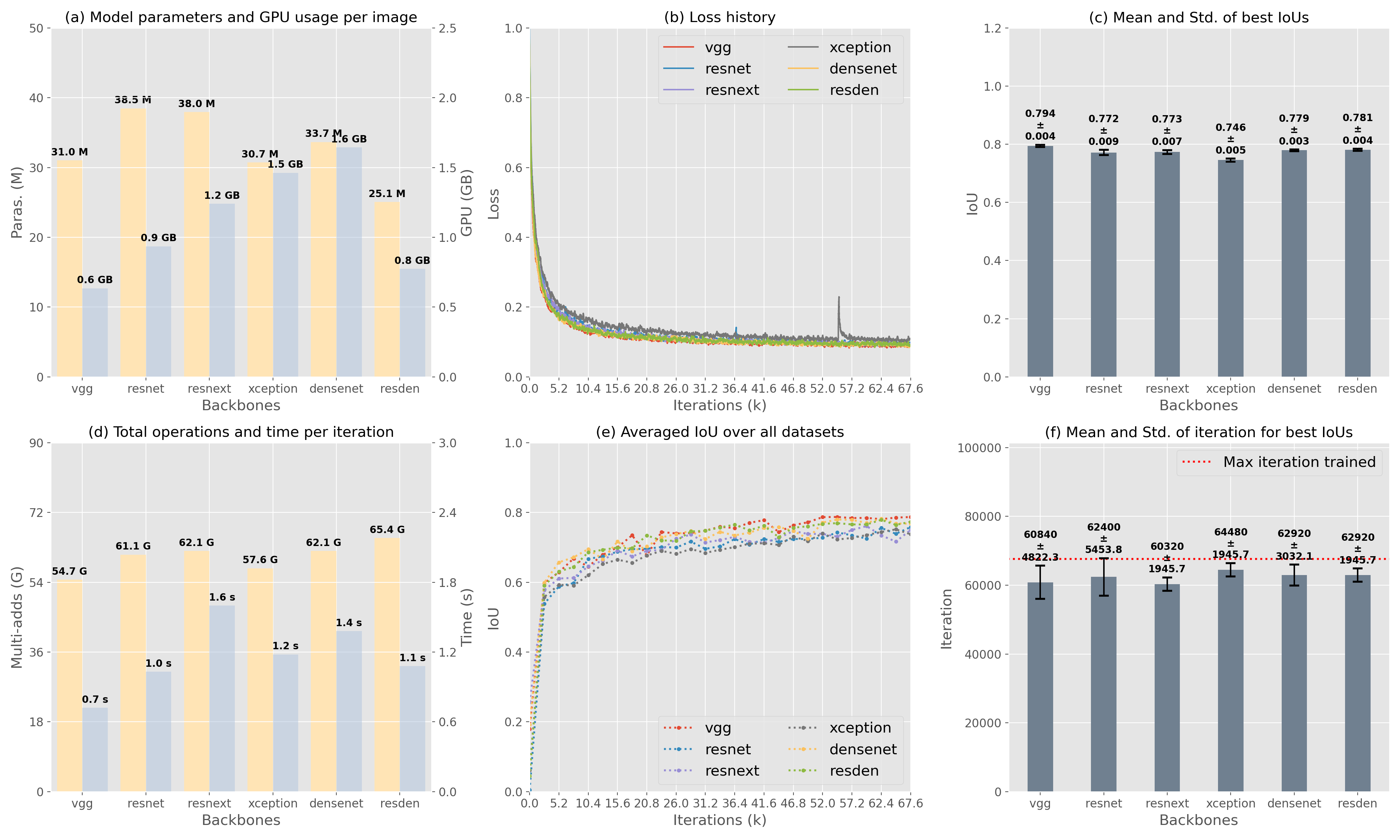}
  \vspace*{-1mm}
  \caption{Results of experiments on encoder backbones. The model parameters in (a) are estimated using Python package `ptflops', and the total operations in (d) are based on a single input of 256$\times$256 pixels. The GPU memory consumption and time per iteration during training are estimated on a single GPU (Tesla P100-PCIE-16GB). The loss history in (b) is smoothed by a 130-point moving average window for better visual presentation. }
  \label{Figure: backbones}
\end{figure*}

\subsection{Effect of backbone}
\label{sec: backbone}
\subsubsection{Experiment details}

Experiments in this section aim to test the performance of the generalized U-Net after replacing its encoder with backbones from other networks. More specifically, the variants evaluated include:

\begin{enumerate} 

\item \textbf{VGG:} The original U-Net with VGG-like \cite{Simonyan2015VeryRecognition} building blocks, and the number of feature channels doubles whenever the spatial dimensions half. All training variants in the previous two sections were trained using this architecture.

\item \textbf{ResNet:} First developed for image classification tasks, the ResNet \cite{He2016DeepRecognition} was also proved to be an effective general feature extractor for other computer vision tasks \cite{He2017MaskR-CNN,Guler2018DensePose:Wild,Lim2017EnhancedSuper-Resolution}, thanks to its powerful residual mappings. We used the ResNet-50 variant since its number of parameters is closest to the VGG backbone.  The `conv1' layer is modified to have a stride of 1, and the bridges are added after each `convX' stage.

\item \textbf{ResNeXt:} The ResNeXt \cite{Xie2017AggregatedNetworks} is an improved version of the ResNet, with aggregated residual transformations implemented as grouped convolutions. Without increasing model complexity, the ResNeXt has better performance than ResNet in image classification and object detection tasks. Again we used the ResNeXt-50 (32$\times$4d) variant for a fair comparison. Similarly, the `conv1' layer is modified to have a stride of 1, and the bridges are added after each `convX' stage. 

\item \textbf{Xception:} The Xception backbone was designed with depthwise separable convolutions as an extreme version of the Inception family \cite{Chollet2017Xception:Convolutions}. It was explored for various computer vision tasks including object detection \cite{Howard2017MobileNets:Applications} and semantic segmentation \cite{Chen2018Encoder-DecoderSegmentation}. Here we only used 4 blocks of the middle flow, such that the resulting model has similar complexity as the VGG baseline.  We also modified the first `Conv 32' layer to have a stride of 1. The bridges are added after the ReLU layers where the feature maps from the main branch and the `Conv $1\times1$' branch merge.

\item \textbf{DenseNet:} The DenseNet was another architecture designed for image classification tasks \cite{Huang2017DenselyNetworks}. It encourages feature reuse and alleviates the vanishing-gradient problem by a densely-connected topological design, without the skip connections as in the previous networks. We used the DenseNet-201 backbone for a fair comparison, which only has slightly more parameters than the VGG baseline.  We modified the first `Conv $7\times7$' layer to have a stride of 1, and the bridges are added within each transition layer before the $2\times2$ average pooling.

\item \textbf{ResDen:} The residual dense network was originally designed for image super-resolution tasks \cite{Zhang2018ResidualSuper-Resolution}, with a combination of ResNet and DenseNet topologies. We set the growth rate at each spatial resolution stage to be one eighth of the number of input feature channels to that stage. We added normalization layer between convolutional layer and activation layer, and at each spatial resolution we use 5 Conv-Norm-ReLU blocks before the downsampling layer (convolution with stride of 2), after which we add the bridge towards the corresponding stage in the decoder. Please see Section \ref{sec: xbnetArch} for a detailed description of the building blocks.

\end{enumerate}

All networks are trained with the Base training configuration from the previous section, but with a batch size of 8 in order to fit the largest model `DenseNet' on GPUs with smaller memory (Tesla P100-PCIE-16GB and Tesla V100-PCIE-16GB).  Again all trainings were carried out for 67,600 iterations, and the models were evaluated on validation sets for every 2,600 iterations.

\subsubsection{Experiment results}

There are some interesting findings by comparison among those tested variants. ResNeXt-50 by design has slightly fewer parameters and more FLOPs than the ResNet-50, which still holds for their respective U-Net integrations as in Figure \ref{Figure: backbones} (a) and (d). On the other hand, the ResNeXt variant only has slight improvements over the ResNet variant as shown in Figure \ref{Figure: backbones} (c), which may suggest that our architecture design has not fully exploited their strong feature extraction powers.  

By using only 4 middle flow blocks, the Xception variant has the closest number of parameters and FLOPs as the original U-Net, while delivering the poorest performance among all the variants explored. By contrast, the DeepLabv3+ \cite{Chen2018Encoder-DecoderSegmentation} integration of Xception for segmentation used 16 middle flow blocks, doubling the number in the original Xception. Therefore the reduced model capacity might be one of the reasons for the compromised performance. Moreover, it also has the slowest convergence speed and unstable loss history as shown in Figure \ref{Figure: backbones} (b).

DenseNet and ResDen variants have very similar network topology that encourages feature reusing and facilitates gradient flow, while the ResDen backbone has additional skip connections that bypass the dense blocks. As a result, with much fewer model parameters, half memory consumption, slightly higher FLOPs but faster training speed, the ResDen variant outperforms the DenseNet variant by a small margin. This encourages new directions to look for suitable backbones for cell segmentation, as the ResDen was developed for image restoration tasks \cite{Xiang2021BoostingSuper-Resolution}, rather than image classification tasks as the other backbones.

However, when compared with the VGG baseline, the above variants with more complex feature extractors have more parameters, consume more GPU memories during training as shown in Figure \ref{Figure: backbones} (a), demand more computations and therefore take longer to train as shown in Figure \ref{Figure: backbones} (d). Nevertheless, it is surprising to see that they have inferior final IoU performances when compared to the VGG baseline as shown in Figure \ref{Figure: backbones} (c). We suspect this is due to one or more of the following reasons:

\begin{enumerate} 
    \item \textbf{Suboptimal architecture design:} In current study the encoders are simply replaced by the corresponding network backbones, before their final layers specialized for image classification or restoration tasks, with minor changes. The bridges are also blindly inserted between the encoder and decoder stages with the same spatial resolutions, with only occasional 1$\times$1 convolutions to match the number of feature channels when necessary. Future works may include careful design considerations to look for the best places to insert the bridges, and proper modification of the backbones to fully utilize their feature extraction powers.
    \item \textbf{Biased training settings:} Current training scheme and settings were selected using the VGG baseline, and might not be optimized for the other encoder variants. For example, under current learning rate some of the variants do not seem to be sufficiently trained at the final iteration, as their averaged IoUs in Figure \ref{Figure: backbones} (e) are still on the rise. Repeat experiments in the previous two sections using a particular encoder of interest may help achieve better results.
    \item \textbf{Random initialization:} To give a fair comparison, the network weights for all variants were initialized randomly and trained from scratch, which might not be ideal for some backbones. For example, although some works report that the final performance of the ResNet family is relatively insensitive to choice of initial weights \cite{Taki2017DeepInitialization}, there are also initialization tricks that might speed up their convergence \cite{Goyal2017AccurateHour}. Moreover, it is also possible to initialize the encoder weights using state-of-the-art pre-trained weights on other datasets (e.g. CIFAR-10, ImageNet \cite{Krizhevsky2012LearningImages,Russakovsky2015ImageNetChallenge}), which might further boost the performance of those networks.
\end{enumerate} 

In a separate study we also replaced the decoder with basic building blocks from those networks, and combined with their corresponding encoder variants. But again we did not observe noticeable improvement of IoUs when compared with the VGG baseline. Therefore, for those new network backbones that have been proven to be stronger feature extractors for other computer vision tasks, how to effectively integrate them into the U-Net architecture and then optimize for cell segmentation tasks is still an open question \cite{Gadosey2020SD-UNET:Budgets,Zhang2020ComparisonNetwork,Chen2020ResidualImages,Chu2019Sea-LandCRF}.

\subsection{Effect of individual modules}

\label{sec: modules}

\begin{figure*}[!ht]
  \includegraphics[width=2.0\columnwidth]{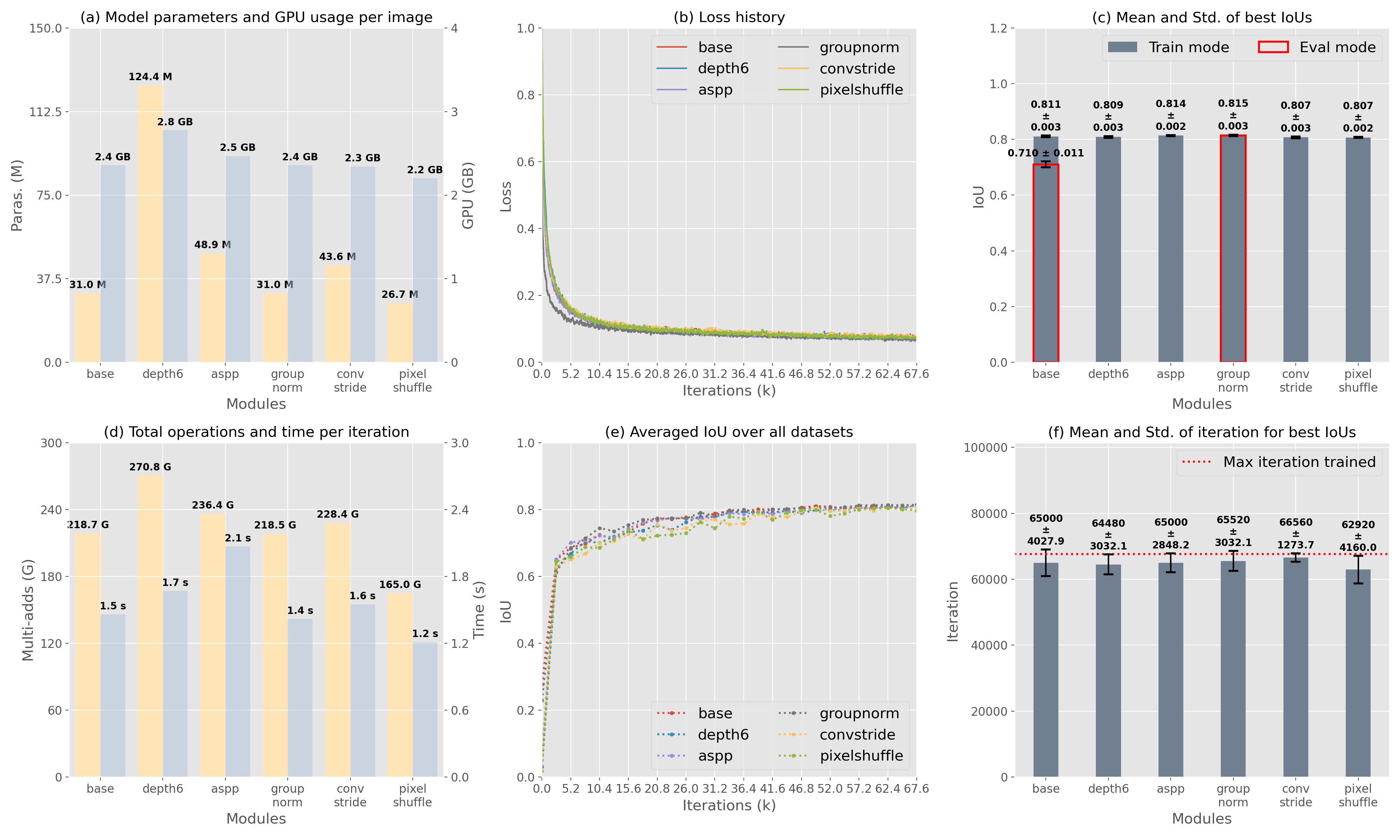}
  \vspace*{-1mm}
  \caption{Results of experiments on individual modules. The model parameters in (a) are estimated using Python package `ptflops', and the total operations in (d) are based on a single input of 512$\times$512 pixels. The GPU memory consumption and time per iteration during training are estimated on a single GPU (Tesla P100-PCIE-16GB). The loss history in (b) is smoothed by a 130-point moving average window for better visual presentation.}
  \label{Figure: modules}
\end{figure*}
\subsubsection{Experiment details}

\begin{figure*}[!b]
  \includegraphics[width=2.0\columnwidth]{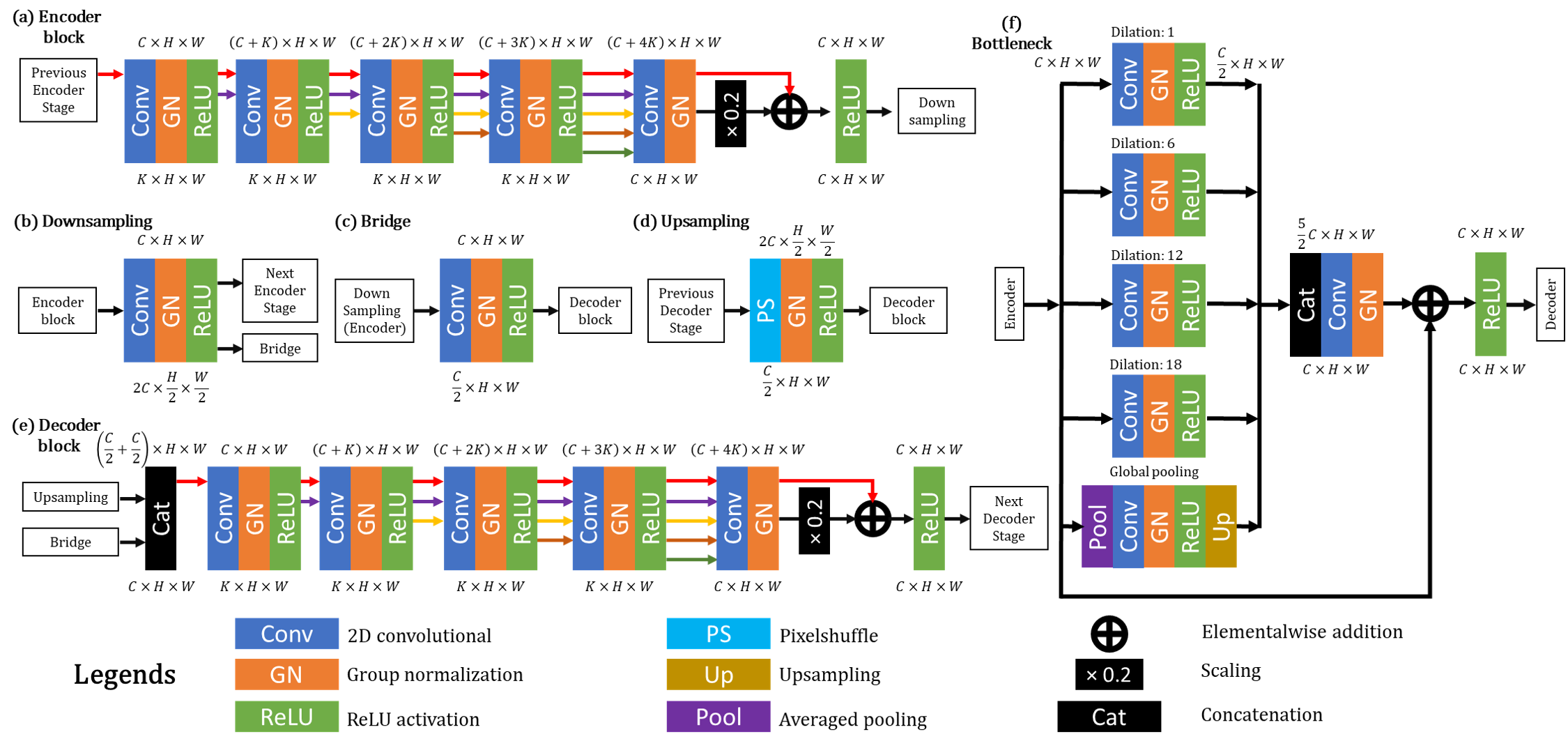}
  \vspace*{-0mm}
  \caption{The network architecture used for the 6th CTC primary track competition. The sizes for input and output feature maps of each unit are shown on the top and bottom of each block, respectively. In the ResDen blocks, we set growth rate $K=C/4$. The usages for (a) encoder block, (b) downsampling block, (c) bridge, (d) upsampling block, (e) decoder block, and the (f) bottleneck are described in the generalized U-Net shown in Figure \ref{Figure: unetArc}.}
  \label{Figure: xbArch}
\end{figure*}

Experiments in this section were carried out by replacing some individual modules in the original U-Net, and the performance of the following variants was evaluated:

\begin{enumerate} 

\item \textbf{Base:} The baseline U-Net architecture, with 5 stages of VGG-like blocks (4 decoder blocks + bottleneck), maxpooling in the encoder, upconvolution in the decoder, batch normalization layers, and no ASPP module (see below) at the bottleneck. All following variants are designed to change only one module with respect to the baseline.

\item \textbf{Depth6:} After adding one more stage to the encoder and decoder, each 1$\times$512$\times$512 input image crop will generate feature maps of 2048$\times$16$\times$16 at the bottleneck.

\item \textbf{Convstride:} We replaced the 2$\times$2 maxpooling layers in the encoder with 3$\times$3 convolutional layers with stride of 2 \cite{Springenberg2015StrivingNet}, which allows learnable downsampling of the feature maps.

\item \textbf{ASPP:} To arbitrarily control the receptive fields of the filters and extract contextual information, we construct the Atrous Spatial Pyramid Pooling \cite{Chen2017RethinkingSegmentation} module with 5 pathways (dilation rates of 1, 6, 12, 18, and a global pooling, as in Figure \ref{Figure: xbArch}f), where each pathway has 512 feature maps. The ASPP is inserted at the bottleneck, with input and output of equally 1024 feature maps.

\item \textbf{Pixelshuffle:} The pixelshuffle layer \cite{Shi2016Real-TimeNetwork} is an efficient sub-pixel convolution layer that shuffles each $2C\times H/2 \times W/2$ feature map from the previous decoder stage to its $C/2\times H \times W$ upsampled version with the same number of elements, which is then concatenated with feature maps of $C\times H \times W$ from the corresponding encoder stage. Next, the VGG block in current decoder stage now takes in $3/2C$ feature maps and produces the same $C$ feature maps as the baseline model.

\item \textbf{Groupnorm:} The group normalization layer \cite{Wu2018GroupNormalization} replaces the batch normalization layer, with the aim to reduce inaccurate batch statistics estimation caused by small batch size. It is especially helpful for segmentation tasks as more contextual information helps scene understanding via larger image crop, which naturally leads to smaller batch sizes. We use a universal group number of 8 for this layer throughout the network.

\end{enumerate}

All networks in this section are trained with the Crop training configuration from Section \ref{sec:trainSettings}, with 512$\times$512 random crops and a batch size of 5 in order to fit the largest model `Depth6' onto GPUs with smaller memory (Tesla P100-PCIE-16GB and Tesla V100-PCIE-16GB).

\subsubsection{Experiment results}

We report the number of model parameters and GPU memory usage per 512$\times$512 image crop during training in Figure \ref{Figure: modules} (a), and total multiply-add operations and time required for forward/backward pass for a minibatch in Figure \ref{Figure: modules} (d). The best IoUs on validation sets and their corresponding iteration numbers are shown in Figure \ref{Figure: modules} (c) and (f).

For the Depth6 variant, adding one more stage in the U-Net significantly increases the model complexity and computational cost, but interestingly it has adverse effect on the best IoU when compared with the baseline model in Figure \ref{Figure: modules} (c). This shows that simply adding more parameters to CNN will not guarantee performance improvement, and abstracting the cell images to $1/32\times1/32$ of its original size does not provide additional semantic information useful for segmentation.

The Aspp variant also has more parameters than the baseline model, and it requires the longest time to carry out a training pass, but it can slightly improve the IoU over the baseline. This improvement could come from the fact that ASPP, when inserted at the bottleneck where the semantic information is the richest, can help understand the contexts for better segmentation. But it could also be a product of increased model complexity (added 17.8 M parameters and 17.7 MAG operations over the baseline model). More experiments with parameterized ASPP design and controlled number of model parameters may help elaborate this in further details.

The Groupnorm variant has exactly the same model complexity and GPU memory consumption as the baseline model, with slightly fewer operations and shorter training time. However, this simple replacement of the normalization layer gives the largest improvement of IoU among all tested variants. Moreover, we reveal another difference between batchnorm and groupnorm layers by inferencing the network in two different modes: training mode and evaluation mode. As demonstrated with the baseline model in Figure \ref{Figure: modules} (c), we empirically found that inferencing in training mode, i.e. normalizing the feature maps using mean and standard deviation estimated from different datasets throughout the training process, leads to inferior results when compared with inferencing in training mode, where instead the minibatch statistics are used for normalization. 

We suspect this comes from the fact that the datasets included in current study are extremely diverse. In our training process with gradient accumulation, the samples from different datasets are never mixed during single forward/backward passes. As a result, the network always sees samples from each single dataset during training, just like in the inferencing stage. On the other hand, with groupnorm layer, inferencing in training model and evaluation mode give equivalent IoUs, indicating that statistics estimated within channel groups are more reliable than along the batch dimension, and more forgiving about training configurations. Another advantage that the groupnorm layer gives is the faster convergence speed. Although all tested models give more or less comparable results after the entire 67.6k training iterations, the Groupnorm variant shows the fastest loss drop and slightly better averaged IoU than other variants with batchnorm during the initial training stage, as shown in Figure \ref{Figure: modules} (b) and (e), which gives another evidence that statistics estimated within channel groups are more effective for normalization in the context of cell segmentation. Further studies could experiment with the number of groups for each stage, and evaluate other types of normalization layers (\cite{Salimans2016WeightNetworks,Qiao2019Micro-BatchStandardization,Ulyanov2016InstanceStylization,Ba2016LayerNormalization}) as well.

For modifications in terms of downsampling in the encoder and upsampling in the decoder, the Convstride and Pixelshuffle variants both show lower IoUs than the baseline network. However, the former has more parameters and requires more operations during training, while the latter is more light-weighted thanks to its smaller decoder (7.84 M/96.94 MAG vs 12.19 M/150.66 MAG for parameters/operations of the baseline model). This suggests that the max-pooling layer is simple yet effective, and should be kept in the network design, while the pixelshuffle layer could be an alternative to consider when model complexity or computational budget is under constraint.

\begin{table*}[!t]
\caption{Dataset splitting rules for training and validation sets for all data configurations.}
\centering
\begin{tabular}{ | m{4cm} | m{5cm}| m{5cm} | } 
\Xhline{4\arrayrulewidth}
\makecell{\textbf{Data configuration}} & \makecell{\textbf{Training set}} & \makecell{\textbf{Validation set}} \\ 
\Xhline{4\arrayrulewidth}
\makecell{GT only} & \makecell{90\% of GT frames for each dataset}  & \makecell{10\% of GT frames 
for each dataset} \\
\hline
\makecell{ST only} & \makecell{90\% of ST frames for each dataset}  & \makecell{10\% of ST frames for each dataset} \\
\hline
\makecell{GT + ST} & \makecell{All ST frames for each dataset}  & \makecell{All GT frames for each dataset} \\
\Xhline{4\arrayrulewidth}
\makecell{all GT} & \makecell{90\% of GT frames for all datasets}  & \makecell{10\% of GT frames 
for all datasets} \\
\hline
\makecell{all ST} & \makecell{90\% of ST frames for all datasets}  & \makecell{10\% of ST frames for all datasets} \\
\hline
\makecell{all GT + all ST} & \makecell{All ST frames for all datasets}  & \makecell{All GT frames for all datasets} \\
\Xhline{4\arrayrulewidth}
\end{tabular}
\label{tb:data_split}
\end{table*}

\section{Participation in the 6th CTC}
\label{sec:ctc}
\subsection{Introduction of the challenge}
\label{sec:ctc_intro}

The 6th Cell Tracking Challenge (CTC) was organized as part of the ISBI 2021. In this edition, the primary focus is put on methods that exhibit better generalizability and work across most datasets, instead of developing methods optimized for one or a few datasets only. 

For the primary track of the challenge, it is required to test the performance of the algorithm under 6 different data configurations: GT only, ST only, GT + ST, all GT, all ST, all GT + all ST. The former 3 configurations (referred to as `individual training' thereafter) require training separate networks for each dataset, while the latter 3 configurations aim to train 3 universal networks for all 13 datasets (referred to as `universal training' thereafter). 

In the original submission before the ISBI 2021, our method ranked as the best runner up among all participants. Later in the additional competition held in June 2021, we also secured the 3rd place with a refined submission.

\subsection{Dataset preparation}

For each data configuration, the training and validation sets were split according to the rules in Table \ref{tb:data_split}. The normalized images and tertiary training targets are prepared as described in Section \ref{sec:imgPre} and Section \ref{sec:trainTarget}.  We use one dataloader per dataset, and the customized dataloader class can yield infinite data stream by resetting its iterator when samples are exhausted. For each dataloader, we apply random shuffle, data augmentation, and random cropping as discussed in Sec \ref{sec: implementation}. Based on the discovery in Section \ref{sec:trainSettings}, the window sizes for random cropping are set to the short side length of the original images in each dataset to aggregate contextual information and gather global features. The batchsize is then set individually for each dataset to the maximum possible size on a 16 GB GPU. 

\subsection{Network architecture}
\label{sec: xbnetArch}

With limited computational resources, in order to complete all 42 ($=13\times3+3$) required trainings for the primary track before the submission deadline, we modified the original U-Net architecture in the following aspect:

\begin{table*}[!t]
\caption{Training settings for different data configurations.}
\centering
\begin{tabular}{ | m{4cm} | m{6cm}| m{6cm} | } 
\Xhline{4\arrayrulewidth}
\makecell{\textbf{Training scheme}} & \makecell{\textbf{Individual training}} & \makecell{\textbf{Universal training}} \\ 
\hline
\makecell{Data configuration} & \makecell{GT only, ST only, GT + ST} & \makecell{all GT, all ST, all GT + all ST} \\ 
\Xhline{4\arrayrulewidth}
\makecell{$\var{dataloaders}$} & \makecell{Only one dataset}  & \makecell{One dataloader per dataset} \\
\hline
\makecell{$\var{model}$} & \makecell{One model per dataset}  & \makecell{One model for all datasets} \\
\hline
\makecell{One epoch} & \makecell{all minibatches from one dataset}  & \makecell{20 minibatches from each dataset} \\
\hline
\makecell{$itr_{epoch}$} & \makecell{(Number of samples) // batchsize}  & \makecell{20$\times$(Number of datasets)} \\
\hline
\makecell{$epoch_{max}$} & \makecell{2D: 1500; 3D: 100}  & \makecell{1500} \\
\hline
\makecell{\rule{0pt}{2.5ex}Scheduler profile} & \makecell{2D: Cosine annealing with warm restarts\\3D: Cosine annealing}  & \makecell{\rule{0pt}{2.5ex} Cosine annealing with warm restarts} \\
\hline
\Xhline{4\arrayrulewidth}
\end{tabular}
\label{tb:training_setting}
\end{table*} 

\begin{enumerate} 
\item We built the network backbone for both the encoder and decoder using the ResDen basic building block as introduced in Section \ref{sec: backbone}.
\item We used convolution with stride of 2 for downsampling, and pixelshuffle for upsampling. $1\times1$ convolutions are used in the bridges to match the channels of feature maps.
\item We inserted an ASPP module with skip connection at the bottleneck, and replaced all batchnorm layers with groupnorm (GN) layers throughout the network as introduced in Section \ref{sec: modules}.
\item Before the first encoder block, we also added an entry flow block (simple Conv-GN-ReLU-Conv-GN-ReLU) that produces the feature maps with the same spatial resolution as the input image to pass through the bridge towards the final decoder block.
\item The first stage of the network now produces only 32 feature maps, halving the numbers in the original U-Net. In the deeper stages, the number of feature channels doubles each time the spatial dimensions half, following the same rule as in the original U-Net.
\end{enumerate}

The resulting network (XpressBio Net) as shown in Figure \ref{Figure: xbArch} is light-weighted with only 11.38 M parameters and 78.18 GMac operations for a 512 $\times$ 512 input crop. Our official implementation is publicly available at \url{https://github.com/westgate458/XB-Net}. Please note that, this network was designed specifically for the CTC competition, and it does not necessarily achieve better or comparable performance as the VGG-version U-Net, due to its reduced model complexity and capacity.

\subsection{Model training}

\subsubsection{Training scheme and settings}

All networks were trained following the training scheme Acc from Section \ref{sec:trainScheme} with settings similar to the Prop variant from Section \ref{sec:trainSettings}. Unequally weighted cross-entropy loss and additional J-regularization loss are jointly optimized by the AdamW optimizer as described in Section \ref{sec: implementation}. Gradients from all datasets are accumulated before an optimization step is performed. Detailed settings, definition of epochs, training iterations, and learning rate schedulers are listed in Table \ref{tb:training_setting}, for individual trainings and universal trainings.  

The learning rate scheduler follows the cosine annealing profile \cite{Loshchilov2016SGDR:Restarts} with maximum and minimum learning rates set to $10^{-4}$ and $10^{-6}$, respectively. When warm restarts are enabled, the learning rate resets at epochs of 100, 300, and 700, and training terminates at epoch 1500, after the model has been trained on $itr_{max}$ number of minibatches. The only exception is individual trainings on 3D datasets,  where we only perform 100 epochs and the learning rate scheduler does not admit restarts, due to limited computational resources. For every 10th epoch, we evaluate the segmentation score, given by the Intersection over Union (IoU) metric, on both the training sets and the validation sets using the official evaluation executable as described in Section \ref{sec: implementation}. The model with the highest segmentation score on the validation set is saved as the best model, and used for later inference on test sets.

The procedure for model training is described in pseudo code as Algorithm \ref{alg:training_procedure}, and the maximum number of iterations on Line 2 is calculated by $itr_{max}=epoch_{max} \times itr_{epoch}$.  For individual training on each dataset, only one dataloader is needed, and the for loop on Line 4 only executes once, therefore the optimizer step is performed for each minibatch. The model selection is based on the IoU score on a single validation set. For universal training on all datasets, the Acc training scheme makes sure at each optimizer step the model is optimized for all datasets. On Line 4 minibatches are drawn from all 13 dataloaders in turns, and the optimizer step is performed based on the accumulated gradients. We only draw 20 minibatches from each dataset during one training epoch, and the best model is selected based on the averaged IoU on 13 validation sets.

\begin{figure}[!ht]
  \includegraphics[width=1.0\columnwidth]{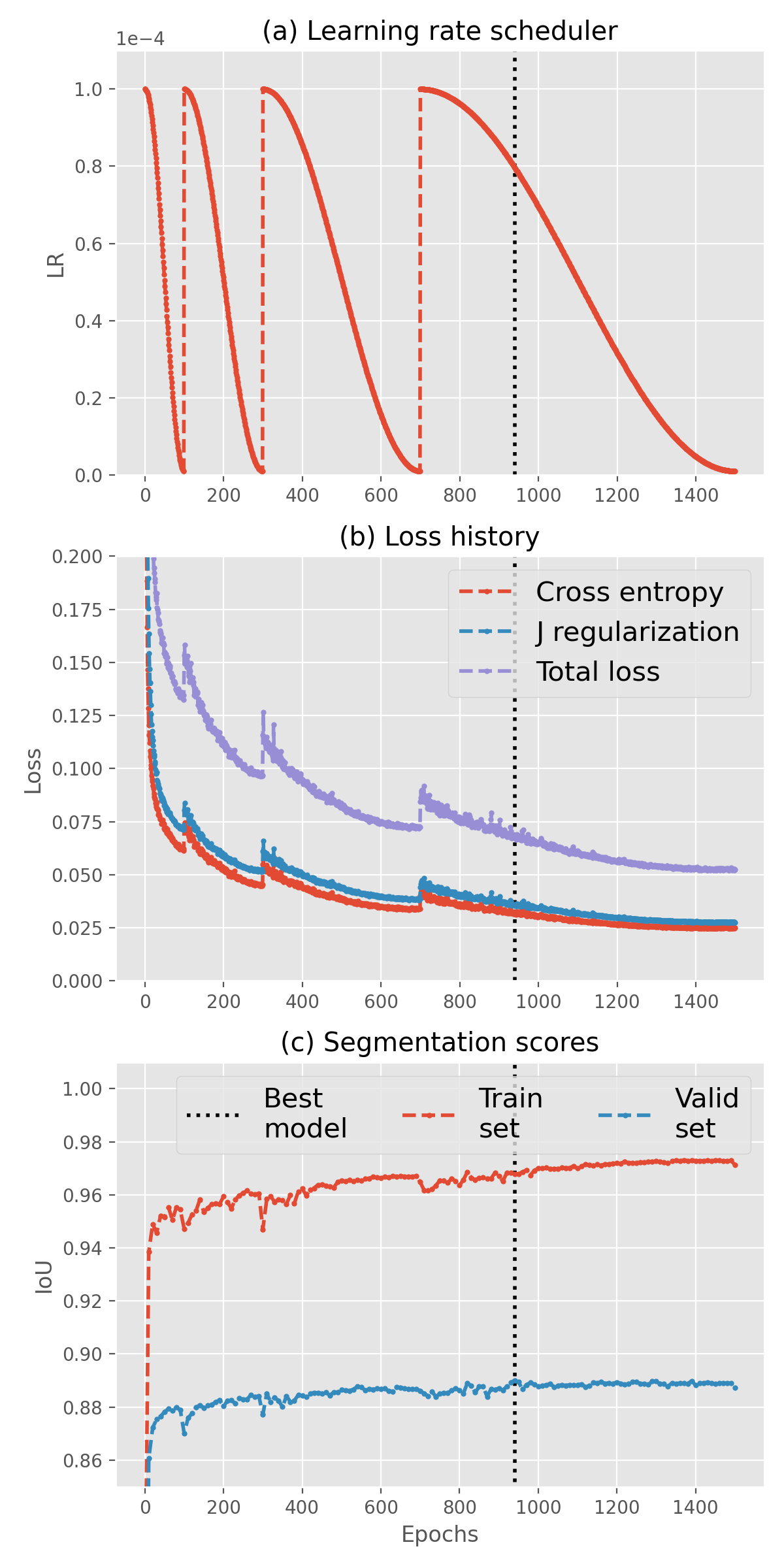}
  \vspace*{-8mm}
  \caption{(a) Learning rate scheduler following the cosine annealing profile with warm restarts. (b) Histories for cross-entropy loss, J regularization loss, and the total loss as the sum of the two. (c) Segmentation metrics calculated by the official evaluation tool. The vertical black dashed lines indicate the best model selected based on the validation IoU score.}
  \label{Figure: lossHistory}
\end{figure}

Representative profiles for learning rate, loss history, and evaluation of IoUs are shown in Figure \ref{Figure: lossHistory}. Please note that, the IoU score on the training set keeps rising as training goes on, while the best model (indicated by the vertical black dashed lines in all three subplots) was selected based on the highest IoU on the validation set shortly after the 3rd restart of the learning rate scheduler. After this point, the model may start suffering from overfitting on the training sets.

\subsection{Inference and post processing}

Inference of the trained models on normalized test images gives tertiary score maps of the original image size. At each pixel location there are three class scores indicating the confidence level that this pixel belongs to one of the three semantic groups:  \verb|background|, \verb|boundary|, and \verb|cell|. We simply take the class with the highest score among all three to be the predicted semantic class of this pixel. 

For 3D datasets, planar segmentation results are converted back to volume stacks first. For all pixels/voxels with label \verb|cell|, we use connected component analysis to group them into distinctly numbered instances.  Since they are naturally separated by the \verb|boundary| class, there is no need for additional splitting operations as in traditional methods and deep learning models trained with binary targets.

The only post-processing steps needed in our algorithm include removing cells that are too small (false alarms), and filling small holes (false negatives) within cell instances. The area thresholds for those operations were determined individually for each dataset, by a coarse grid search on the training sets using the official evaluation tools for detection, segmentation, and tracking scores. 

\subsection{Cell tracking}

After post-processing, the segmented cell instances are approximated by a convex hull and an equivalent ellipse using popular libraries (for example scikit-image \cite{VanDerWalt2014Scikit-image:Python}). Morphological characteristics of those two approximations (for example, max intensity, aspect ratio, solidity, equivalent radius, among others \cite{Rosin2005COMPUTINGMEASURES}) can be used together with the intensity-weighted centroids to represent the cells as abstracted vectors in a higher-dimensional space. Those additional characteristics help distinguish cells when they are over-crowded or the per-frame cell displacements are too large for the simple nearest-neighbor tracking method to draw correct correspondence \cite{Jun2021Multi-parametricPreparation}.

The abstracted cells between frames are then matched by the multi-parametric tracking method \cite{Guo2019MicroscaleVelocimetry} originally designed for particle tracking velocimety (PTV). This method has been used for tracking fibronectin-expressing mesenchymal tumor cells \cite{Jun2020Fibronectin-ExpressingMetastasis}. In the matching process, we manually exclude cells in the second frame that are more than a certain distance away from the cell in the previous frame.  This threshold was chosen empirically by a coarse search on each training set using the official evaluation tool for tracking score. Tracks terminate or originate when there are no matched cells in the next or previous frame. Based on the frame-wise matching results, the cell instance segmentation masks are re-numbered by the track IDs over the entire sequence, with no additional modification to the masks.

\section{Conclusion}
\label{sec:conclusion}

In this paper, we shared some of our recent successes and failures in finding the best way to train a universal segmentation network for various cell types and imaging modalities. Those attempts were motivated by our own previous works in cancer research, as well as by the demands from the cell tracking community. Our current approach was inspired by some recent works on direct instance segmentation (tertiary training target) and novel loss functions (J-regularization). To systematically investigate the effects of training schemes, training settings, network backbones, and individual modules on the cell segmentation performance, we generalized the widely-accepted U-Net architecture, designed a model performance evaluation pipeline, and finally carried out extensive experiments to highlight the importance of each aspect.

Our preliminary findings indicate that the key to training a universal cell segmentation network is all-time supervision on all datasets and sampling each dataset in an unbiased way.  This could be achieved by our proposed training scheme where the minibatches are drawn in turn from each dataset, and the gradients are accumulated before each optimization step. A few training tricks can further boost the IoU performance, including uneven weights in the cross-entropy loss function, well-designed learning rate scheduler, larger image crops for more contextual information, and additional loss terms to pay additional attention to under-presented classes. 

As for individual modules, we found that ASPP module and group normalization layer both can slightly improve the segmentation performance. The former helps semantic understanding by aggregating contextual information at various dilation rates, and the added model parameters and complexity are more rewarding than adding one more stage to the U-Net. Compared with batch normalization, the latter gives more reliable and effective statistics within channel groups for feature map normalization, which is forgiving about training configurations, and also speeds up the initial model convergence.

On the other hand, there are still open questions to answer. For example, our experiments with alternative encoder backbones did not provide us with a satisfying answer regarding how to effectively incorporate those stronger feature extractors into the U-Net architecture. Future studies may include careful modification of those backbones, smarter experiments to find the best layers whose feature maps are best suited for the bridge connections, and evaluation of other novel modules after proper integration into the overall network. Our experiments also suggest that there might exist common features to define cell boundaries across cell types and imaging modalities. More evaluations could be done to test the performance of trained models on totally unseen datasets.

With knowledge gained from this exercise, we designed a light-weighted U-Net architecture (XpressBio Net) with residual-densenet blocks as the backbone, inserted an ASPP module at the bottleneck for contextual information, and replaced  the original upconv layers and batch normalization layers with pixelshuffle layers and group normalization layers to further boost the performance. The networks trained using the proposed gradient accumulation scheme won the best runner-up in the primary track of the 6th CTC, and finally secured a 3rd position after an additional round of competition in preparation for the summary publication.

\section{Acknowledgments}
\label{sec:acknowledgments}
No funding was received for this study. The authors have no relevant financial or non-financial interests to disclose.

\bibliographystyle{IEEEbib}
\bibliography{references}

\end{document}